\documentclass[a4paper,twoside]{article}

\usepackage{epsfig}
\usepackage{subcaption}
\usepackage{calc}
\usepackage{amssymb}
\usepackage{amstext}
\usepackage{amsmath}
\usepackage{amsthm}
\usepackage{multicol}
\usepackage{pslatex}
\usepackage{apalike}

\usepackage{bm}

\usepackage{SCITEPRESS}     

\begin{document}

\title{Time to Focus: A Comprehensive Benchmark Using Time Series Attribution Methods.}


\author{\authorname{Dominique Mercier\sup{1,2}\orcidAuthor{0000-0001-8817-2744}, Jwalin Bhatt\sup{2}, Andreas Dengel\sup{1,2}\orcidAuthor{0000-0002-6100-8255}, and Sheraz Ahmed\sup{1}\orcidAuthor{0000-0002-4239-6520}}
\affiliation{\sup{1}German Research Center for Artificial Intelligence GmbH (DFKI), Kaiserslautern, Germany}
\email{firstname.lastname@dfki.de}
\affiliation{\sup{2}Technical University Kaiserslautern (TUK), Kaiserslautern, Germany}
\email{lastname@rhrk.uni-kl.de}
}

\keywords{Deep Learning, Time series, Interpretability, Attribution, Benchmarking, Convolutional Neural Network, Artificial Intelligence, Survey.}

\abstract{In the last decade neural network have made huge impact both in industry and research due to their ability to extract meaningful features from imprecise or complex data, and by achieving super human performance in several domains. However, due to the lack of transparency the use of these networks is hampered in the areas with safety critical areas. In safety-critical areas, this is necessary by law. Recently several methods have been proposed to uncover this black box by providing interpreation of predictions made by these models. The paper focuses on time series analysis and benchmark several state-of-the-art attribution methods which compute explanations for convolutional classifiers. The presented experiments involve gradient-based and perturbation-based attribution methods. A detailed analysis shows that perturbation-based approaches are superior concerning the Sensitivity and occlusion game. These methods tend to produce explanations with higher continuity. Contrarily, the gradient-based techniques are superb in runtime and Infidelity. In addition, a validation the dependence of the methods on the trained model, feasible application domains, and individual characteristics is attached. The findings accentuate that choosing the best-suited attribution method is strongly correlated with the desired use case. Neither category of attribution methods nor a single approach has shown outstanding performance across all aspects.}

\onecolumn \maketitle

\section{\uppercase{Introduction}}
\label{sec:introduction}
For several years, the field of artificial intelligence has shown a growing interest in both research and industry~\cite{allam2019big}. This attention led to the discovery of crucial limitations and weaknesses when dealing with artificial intelligence. The following main concerns have become increasingly important: resource management, efficiency, data security, but also interpretability and explainability. According to~\cite{perc2019social} these limitations originate from the social and the juristic domain.

Particularly the interpretability of the classifiers' decisions plays a crucial role in industry and safety-critical application areas. The legal situation reinforces the significance of interpretability. In the medical sector, financial domain, and other safety-critical areas~\cite{bibal2020impact} explainable computations are required.

Over several years, a wide range of methods to explain neural networks was summarized by~\cite{dovsilovic2018explainable}. These methods involve both intrinsic and post-hoc approaches across a broad scope of modalities involving language processing, image classification, and time series analysis. The majority of these approaches have origin from image analysis since the visual criteria~\cite{zhang2018visual} and concepts are more intuitive for humans.

Due to the lack of evaluations of the existing approaches in the context of time series, the paper concentrates on their applicability and effectiveness in time series analysis. A comprehensive analysis of existing attribution methods as one class of commonly used interpretability methods is presented. The paper further covers the strengths and weaknesses of these methods. Specifically, a runtime analysis is done, which is relevant for real-time use cases. Besides the computational aspects, the Infidelity, Sensitivity, influence on accuracy, and correlations between the attributions were evaluated. For this purpose, AlexNet was used as architecture and experiments on well-known and freely available time series datasets were executed.

The contribution includes a comprehensive analysis of several state-of-the-art attribution methods concerning runtime, accuracy, robustness, Infidelity, Sensitivity, model parameter dependence, label dependence, and dataset dependence. The findings illustrate the superior performance of gradient-based methods concerning runtime and Infidelity. In contrast,  perturbation-based approaches give better results concerning the Sensitivity, occlusion game, and continuity of the attribution maps. The paper emphasizes that none of the two categories is superior in all evaluated characteristics and that the selection of the best-suited attribution methods depends on the desired properties of the use case.

\section{\uppercase{Related Work}}
\label{sec:related}
Often Attribution methods are used to interpret classifiers. A comprehensive overview of the different categories involving attribution methods is given by Das et al.~\cite{das2020opportunities}. Attribution methods are well-known as they are compatible with various networks and therefore do not require any restrictions in the design of the network. Attribution methods belong to the class of posterior techniques that require less cognitive effort to interpret due to their simple visualization of the relevance of the input. Furthermore, no detailed knowledge about the analyzed classifier is needed. Especially for image classification, there is a wide range of attribution methods and different benchmark works. According to the authors of~\cite{abdul2020cogam}, an explanation always results in a trade-off between accuracy, simplicity, and cognitive effort that is one reason for the popularity of the attribution methods.

Aspects like the Sensitivity, the change of the attribution map by permutation of the input signal, and other metrics are applied to understand the exact advantages and disadvantages of the methods. More details about the importance and impact of Sensitivity are summarized by Ancona et al.~\cite{ancona2017towards}. Besides Sensitivity, Infidelity, known as the change in classification when permutating the input, plays a role. According to Yeh et al.~\cite{yeh2019fidelity}, Infidelity serves a pivotal role in explaining the quality of an attribution method. Further aspects are the runtime and the difference between black box and white box requirements.

Also, aspects like the dependency on gradient calculation play a big role. Some methods work without backpropagation and use permutations and the forward pass to calculate the relevance of the input points. A detailed differentiation of these categories can be was provided by Anacona et al.~\cite{ancona2019gradient} and Ivanovs et al.~\cite{ivanovs2021perturbation}.

The experiments are aligned with existing image processing surveys and used similar metrics. A comprehensive analysis for the image modalities was written by Adebayo et al.~\cite{adebayo2018sanity}. Although this paper used similar experiment settings, the results may differ due to the diverse modalities. 

However, the precise evaluation of these methods in the time series domain is crucial. Karliuk mentioned that~\cite{karliuk2018ethical} it was legally stipulated that neuronal networks, for example, may not be used in all areas of life as their interpretability and ethical problems still exist. Peres et al.~\cite{peres2020industrial} discussed which aspects are relevant for the application of neural networks in the economy. In addition to data protection restrictions and efficiency, the interpretability of neural networks plays a pivotal role, especially today.

\section{\uppercase{Evaluated Methods}}
\label{sec:methods}
This section provides an overview of the different methods, their applicability, and categorization. First of all, the used methods are a subset that can be used in the field of time series analysis and do not require the selection of internal layers for calculation. 

\subsection{Gradient-based}
Gradient-based methods include Integrated Gradients, Saliency maps, InputXGradient, GradientShap~\cite{lundberg2017unified} and Guided-Backprop. In the case of Integrated Gradients~\cite{sundararajan2017axiomatic} backpropagation is applied to calculate an importance value for each input value relative to a baseline. An elementary part of this method is to know the baseline. The selection of this baseline is crucial for the computation of the gradients to make sense. In contrast, the Saliency~\cite{simonyan2013deep} does not need a baseline and only computes the gradients. A method that is very similar to this is called Input X Gradient~\cite{shrikumar2016not}.
Here the calculated gradients are multiplied by the input to create a relation between them and the input values. Guided-Backpropagation~\cite{springenberg2014striving} also uses a backward run to compute the importance of the values. However, a modification to the network is required. The resulting limitation is the access to the activation function to modify it. Previously mentioned methods require a backward calculation leading to noisy explanations due to the gradients. In addition, they need to access internal parameters. The core concept of GradientShap relies on the estimation of the SHAP values of the input.
SHAP values are estimated using targeted permutations of the input sequence. These values are an approximation since the exact calculation of the SHAP values is very time and resource-intensive. GradientShap is in this respect very similar to Integrated Gradients. 

\subsection{Perturbation-based}
These methods are different to the gradient based methods, as they do not need access to the gradients. Perturbation-based methods slightly change the input and compare the output to the baseline to create an importance ranking. Example approaches for this category are Occlusion~\cite{zeiler2014visualizing} and Feature Permutation~\cite{fisher2019all} and FeatueAblation. All these methods differ in the way they modify the individual points. Another method that makes use of the perturbation principle is Dynamask~\cite{Crabbe2021Dynamask}. A mask gets learned utilizing permutations to calculate the relevant input values. Apart from Dynamask, the above methods have the advantage that no backpropagation and thus no full access to the network and the parameters is required. Dynamask particularly allows easy visualization and restriction to a percentage of the features. The disadvantages of these methods are the correct choice of permutation depending on the dataset. In addition, the increased runtimes due to the multiple forward passes are negative too.

\subsection{Miscellaneous}
Shapley Value Sampling (SVS)~\cite{mitchell2021sampling} is based solely on a random permutation of the input values. The influence on the output is determined utilizing multiple forward calculations. Using SVS requires further points in addition to the data point under consideration to be changed. Finally, Lime~\cite{lime} tries to explain the model using a local model trained on perturbed input samples related to the original input to train an interpretable model and create importance values based on this model.

\section{\uppercase{Datasets}}
\label{sec:datasets}

\begin{table}[!t]
\caption{\textbf{UEA \& UCR Datasets} related to critical infrastructures.} 
\label{tab:datasets}
\centering
\scalebox{0.64}{
\begin{tabular}{l|r|r|r|r|r}
\textbf{Domain \& Dataset} & \textbf{Train} & \textbf{Test} & \textbf{Steps} & \textbf{Channels} & \textbf{Classes} \\
\hline\hline
\textbf{Communications} & & & & & \\
UWaveGestureLibraryAll  & $896$     & $3,582$   & $945$ & $1$   & $8$ \\
\hline
\textbf{Critical manufacturing} & & & & & \\
FordA                   & $3,601$   & $1,320$   & $500$ & $1$   & $2$ \\
Anomaly                 & $35,000$  & $15,000$  & $50$  & $3$   & $2$ \\
\hline
\textbf{Public health} & & & & & \\
ECG5000                 & $500$     & $4,500$   & $140$ & $1$   & $5$ \\
FaceDetection           & $5,890$   & $3,524$   & $62$  & $144$ & $2$ \\
\hline
\textbf{Telecommunications} & & & & & \\
CharacterTrajectories   & $1,422$   & $1,436$   & $182$ & $3$   & $20$ \\
\end{tabular}
}
\end{table}

For the experiments asubset of the datasets from UEA \& UCR~\cite{tsc2021datasets} repositories was used. The selected datasets cover different aspects such as a variance in the number of channels, sequence length, classes, and task. The tasks include point anomaly and sequence anomaly classification in which an occurrence of a single anomalous point is enough to change the label. Furthermore, the datasets cover traditional sequence classification not related to atypical behavior. These datasets are taken from different critical domains that require explainability and in addition privacy. In addition, to the UEA \& UCR datasets, The point anomaly dataset proposed by Siddiqui et al.~\cite{siddiqui2019tsviz} was included as it is unique compared to the others, and a perturbation on single points can change the complete prediction. Table~\ref{tab:datasets} lists the different datasets used in this paper.

\section{\uppercase{Experiments \& Results}}
\label{sec:experiments}
In this section, different aspects of the above methods are evaluated. The methods were not optimized to ensure fairness among the approaches. Fine-tuning an attribution method requires assumptions about the dataset. However, in a real case, this prior knowledge is not necessarily given. The work covers the following aspects: Impact on the accuracy, Infidelity, Sensitivity, runtime, the correlation between the methods, and impact of label and model parameter randomization. In existing work such as~\cite{adebayo2018sanity,huber2021benchmarking,nielsen2021robust} these measurements are judged as significant.

In general, all experiments are executed for the previously mentioned datasets. However, identical results were excluded due to the limited space and the low amount of insights they provide to the reader. The preprocessing of the data covers a standardization to achieve a mean of zero and a standard deviation. Therefore, the baseline signal is a sequence of zeros. AlexNet was modified to work with 1D data and trained the network using an SGD optimizer and a learning rate of $0.01$ to evaluate the different attribution techniques. In Table~\ref{tab:architecture} the network structure of the AlexNet is shown. The layer names used in the reset of the paper refer to those mentioned in the architecture figure. All networks were trained for a maximum of $100$ epochs. In addition, the learning rate was reduced by half after a plateau and performed early stopping based on the validation set. In the particular case of label permutation, the labels of the training data were randomized.  All experiments used fixed random seeds to preserve reproducibility.

Due to the immense computational effort, a set of $100$ test samples was selected to evaluate the attribution methods. In addition, these samples preserve the class distribution of the test set. In Table~\ref{tab:accs} the weighted f1 scores are shown. The differences in the weighted-f1 scores between the original data and the subsets are less than $5\%$. Only the FaceDetection dataset shows a difference of $19\%$. This difference does not hinder the analysis as those two sets are never compared.

\begin{table}[!t]
\caption{\textbf{Architecture.} AlexNet architecture includes layer names used in this paper. Dropout layers are excluded from the table. The padding of every layer was set to 'same'. The variables 'c', 'w', and 'l' depend on the input channels, width, and the number of classes of the used dataset.}
\label{tab:architecture}
\centering
\scalebox{0.74}{
\begin{tabular}{l|l|c|c|c|c}
\textbf{Name} & \textbf{Type} & \textbf{In} & \textbf{Out} & \textbf{Size} & \textbf{Stride} \\
\hline
conv\_1  & Conv, ReLu, Batch & $c$           & $96$      & $11$  & $4$ \\
pool\_1  & MaxPool           & $96$          & $96$      & $3$   & $2$ \\
\hline
conv\_2  & Conv, ReLu, Batch & $96$          & $256$     & $5$   & $1$ \\
pool\_2  & MaxPool           & $256$         & $256$     & $3$   & $2$ \\
\hline
conv\_3  & Conv, Relu, Batch & $256$         & $384$     & $3$   & $1$ \\
conv\_4  & Conv, Relu, Batch & $384$         & $384$     & $1$   & $1$ \\
conv\_5  & Conv, Relu, Batch & $384$         & $256$     & $1$   & $1$ \\
pool\_2  & MaxPool           & $256$         & $256$     & $3$   & $2$ \\
\hline
dense\_1 & Dense, ReLu       & $w * 256$     & $4,096$   &       & \\
dense\_2 & Dense, ReLu       & $4,096$       & $4,096$   &       & \\
dense\_3 & Dense             & $4,096$       & $l$       &       & \\
\end{tabular}
}
\end{table}

\begin{table}[!t]
\caption{\textbf{Accuracies.} Evaluation of the test data using the original split provided by the datasets. Subset covers the performance of the model on the $100$ samples subset that is used in the rest of the paper due to the computational limitations. The values show the weighted-f1 scores and provide evidence that the difficulty of the sets is similar.}
\label{tab:accs}
\centering
\scalebox{0.82}{
\begin{tabular}{l|c|c}
\textbf{Dataset} & \textbf{Test Set} & \textbf{Attribution. Subset} \\
\hline
Anomaly                 & $0.9801$ & $0.9464$ \\
CharacterTrajectories   & $0.9930$ & $1.0000$ \\
ECG5000                 & $0.9352$ & $0.8907$ \\
FaceDetection           & $0.5956$ & $0.7097$ \\
FordA                   & $0.9204$ & $0.9400$ \\
UWaveGestureLibraryAll  & $0.9318$ & $0.9802$ \\
\end{tabular}
}
\end{table}

\subsection{Impact on the Accuracy}
\begin{figure*}[!t]
\centering
\includegraphics[width=\linewidth]{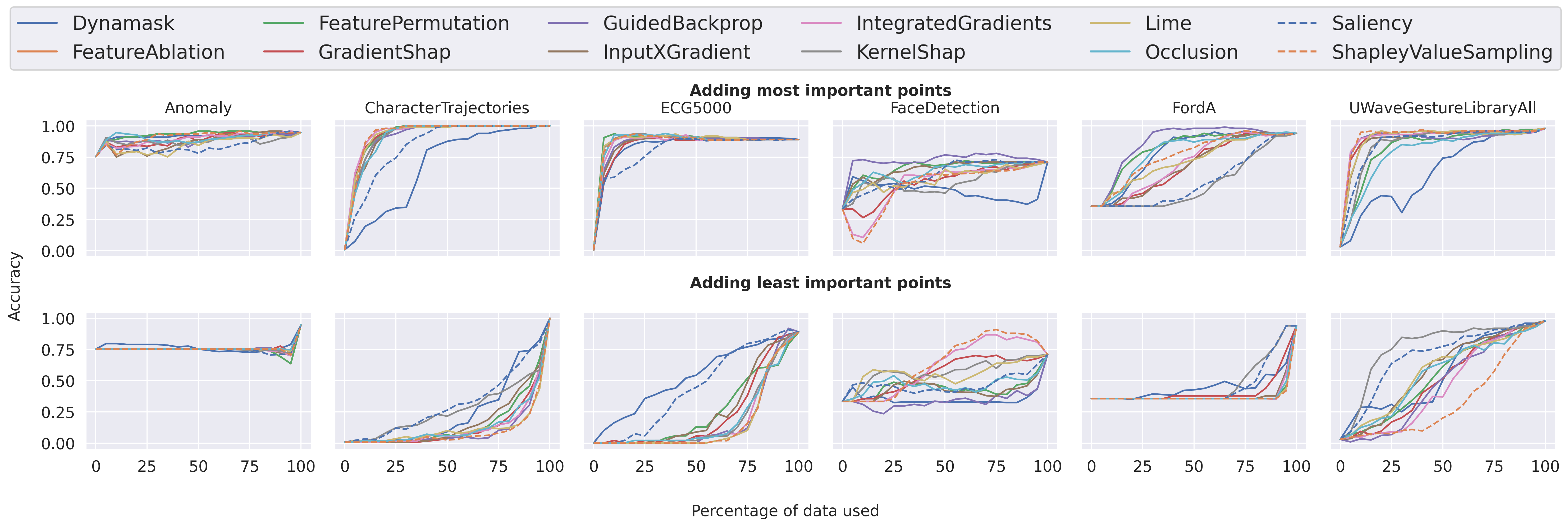}
\caption{\textbf{Impact on accuracy.} Shows the impact when adding points to the baseline signal using the attribution scores as sequence order. \textbf{Top:} Shows the increase adding the most important points. \textbf{Bottom:} Shows the increase adding the least important points. Precisely, each plot read from $100$ percent used data to 0 shows the impact removing the least important points for the top row, respectively the most important for the bottom row. The values show the weighted-f1 scores. Except Dynamask, Saliency, and KernelShap the performances of the approaches are similar.}
\label{fig:accuracy_drops}
\end{figure*}

To evaluate the performance of the attribution methods, the drop in accuracy under the addition and occlusion of the data points was inspected. To occlude the data, the points were set to zero as this is the mean of the data corresponding to the baseline. Respectively, the start point is zero when adding points step-wise. This experiment was performed in both directions adding important points and insignificant data. In Figure~\ref{fig:accuracy_drops} the results show that most of the methods were able to correctly identify the data points that have the most influence on the accuracy. Intuitively, data points that have a higher impact on accuracy should be ranked higher. The top row shows the accuracy increase adding the most significant points step-wise. The bottom row shows the behavior of adding the insignificant data points first. Ultimately, reading each plot starting from $100$ to $0$ percent results in excluding the least important ones for the top row and most important ones for the bottom row. The experiments highlight that for most datasets, namely Anomaly, CharacterTrajectories, ECG5000, and  UWaveGestureLibraryAll, a small number of data points is enough to recover the accuracy. Surprisingly, adding unimportant data points resulted in higher accuracy values. Examples of this behavior are the Lime, Saliency, and Dynamask approach. This behavior appears in the ECG5000, FordA, and UWaveGestureLibraryAll datasets. Saliency has shown to suffer from the noisy backpropagation. The drawbacks of Lime and Dynamask are their hyperparameters. These are the number of neighborhood samples for Lime and the area size and continuity loss for Dynamask.

\begin{table*}[!t]
\caption{\textbf{Prediction agreement.} Evaluation of how many data points are required to reach a specific agreement between the original and modified input. All numbers are in percentage, and lower numbers are better as less data was needed to restore the ground-truth predictions. The numbers in each cell show the percentage of data points added to the baseline to achieve the required agreement concerning the prediction. Perturbation-based approaches have shown a significantly better performance.}
\label{tab:agree}
\centering
\scalebox{0.68}{
\begin{tabular}{l|c|c|c|c|c|c|c|c|c|c|c|c|c|c|c|c|c|c}
\textbf{Method} & \multicolumn{3}{c|}{\textbf{Anomaly}} & \multicolumn{3}{c|}{\textbf{CharacterTraj.}} & \multicolumn{3}{c|}{\textbf{ECG5000}} & \multicolumn{3}{c|}{\textbf{FaceDetection}} & \multicolumn{3}{c|}{\textbf{FordA}} & \multicolumn{3}{c}{\textbf{UWaveGesture}} \\
Req. Agreement in [\%] & $90$ & $95$ & $100$ & $90$ & $95$ & $100$ & $90$ & $95$ & $100$ & $90$ & $95$ & $100$ & $90$ & $95$ & $100$ & $90$ & $95$ & $100$ \\
\hline
\textbf{Gradient-based} & & & & & & & & & & & & & & & & & \\
GradientShap~\cite{lundberg2017unified}
                    & $\bm{1}$ & $44$ & $97$ & $\bm{15}$ & $18$ & $32$ & $15$ & $20$ & $75$ & $60$ 
                    & $71$ & $98$ & $69$ & $77$ & $\bm{96}$ & $\bm{12}$ & $38$ & $100$ \\
GuidedBackprop~\cite{springenberg2014striving}
                    & $\bm{1}$ & $76$ & $98$ & $17$ & $27$ & $45$ & $13$ & $\bm{14}$ & $83$ & $\bm{2}$ 
                    & $\bm{2}$ & $\bm{5}$ & $\bm{33}$ & $61$ & $98$ & $\bm{11}$ & $\bm{16}$ & $100$ \\
InputXGradient~\cite{shrikumar2016not}
                    & $\bm{1}$ & $51$ & $92$ & $16$ & $21$ & $\bm{29}$ & $18$ & $24$ & $\bm{42}$ 
                    & $26$ & $36$ & $55$ & $69$ & $81$ & $98$ & $\bm{12}$ & $38$ & $100$ \\
IntegratedGradients~\cite{sundararajan2017axiomatic}
                    & $\bm{1}$ & $\bm{3}$ & $99$ & $\bm{12}$ & $\bm{15}$ & $31$ & $11$ & $18$ 
                    & $\bm{38}$ & $63$ & $81$ & $97$ & $70$ & $79$ & $98$ & $\bm{12}$ & $39$ & $100$ \\
Saliency~\cite{simonyan2013deep}
                    & $\bm{1}$ & $76$ & $97$ & $34$ & $41$ & $48$ & $32$ & $37$ & $75$ & $48$ & $51$ 
                    & $54$ & $88$ & $93$ & $100$ & $20$ & $53$ & $100$ \\
\hline
\textbf{Perturbation-based} & & & & & & & & & & & & & & & & & \\
Dynamask~\cite{Crabbe2021Dynamask}
                    & $\bm{1}$ & $5$ & $100$ & $55$ & $72$ & $92$ & $18$ & $31$ & $100$ & $100$ 
                    & $100$ & $100$ & $50$ & $71$ & $100$ & $61$ & $74$ & $\bm{98}$ \\
FeatureAblation~\cite{zeiler2014visualizing}
                    & $\bm{1}$ & $\bm{2}$ & $\bm{48}$ & $\bm{15}$ & $20$ & $\bm{28}$ & $\bm{6}$ 
                    & $\bm{9}$ & $60$ & $\bm{25}$ & $\bm{30}$ & $\bm{35}$ & $\bm{44}$ & $\bm{52}$ 
                    & $\bm{82}$ & $26$ & $55$ & $\bm{99}$ \\
FeaturePermutation~\cite{fisher2019all}
                    & $\bm{1}$ & $\bm{2}$ & $\bm{48}$ & $\bm{15}$ & $20$ & $\bm{28}$ & $\bm{6}$ 
                    & $\bm{9}$ & $60$ & $\bm{25}$ & $\bm{30}$ & $\bm{35}$ & $\bm{44}$ & $\bm{52}$ 
                    & $\bm{82}$ & $26$ & $55$ & $\bm{99}$ \\
Occlusion~\cite{zeiler2014visualizing}
                    & $\bm{1}$ & $3$ & $83$ & $19$ & $20$ & $\bm{29}$ & $9$ & $15$ & $\bm{46}$ 
                    & $\bm{16}$ & $47$ & $87$ & $43$ & $\bm{55}$ & $\bm{96}$ & $33$ & $68$ & $100$ \\
\hline
\textbf{Miscellaneous} & & & & & & & & & & & & & & & & & \\
KernelShap~\cite{lundberg2017unified}
                    & $\bm{1}$ & $58$ & $100$ & $\bm{15}$ & $22$ & $43$ & $\bm{8}$ & $15$ & $84$ 
                    & $70$ & $84$ & $99$ & $90$ & $94$ & $98$ & $16$ & $34$ & $100$ \\
Lime~\cite{lime}     
                    & $\bm{1}$ & $90$ & $100$ & $\bm{15}$ & $\bm{17}$ & $49$ & $\bm{8}$ & $17$ & $75$ 
                    & $49$ & $52$ & $81$ & $79$ & $86$ & $99$ & $13$ & $\bm{17}$ & $100$ \\
ShapleyValueSampling~\cite{mitchell2021sampling}
                    & $\bm{1}$ & $30$ & $\bm{51}$ & $\bm{12}$ & $\bm{13}$ & $30$ & $10$ & $18$ & $71$ 
                    & $68$ & $90$ & $93$ & $65$ & $79$ & $97$ & $\bm{9}$ & $\bm{15}$ & $100$ \\
\end{tabular}
}
\end{table*}

\subsection{Prediction Agreement}
In addition to the accuracy drops, the agreement with the original data was computed. Therefore, In Table~\ref{tab:agree} the percentage of data required to produce a similar prediction as with the original sample are shown. To do so, data points are included step-wise based on their importance. Initially, all data samples start with zeros. In every step, the next most important data point was added. The results show that the required data for an agreement of $90\%$ of the predictions is in most cases reached with far less than $50\%$ of the data. The results show that the perturbation-based approaches overall performed better. In addition, the results show that the required amount of data highly differs based on the dataset. Intuitively, Dynamask did not perform well on this task as it provides only a binary decision on whether a feature is significant or not. Besides Dynamask, the Saliency and KernelShap have shown a worse performance too. On the other side, the FeatureAblation, FeaturePermutation, GuidedBackProp, and ShapleyValueSampling approaches have shown superior performance to the other methods using the data suggested to be important by those methods resulted in a much earlier agreement of the prediction. Interestingly, the point anomaly dataset has shown that highlighting only one percent of the data is enough to reach a $90\%$ agreement. In addition, getting to a similar prediction for the UWaveGesture dataset required every method to include almost every point.

\begin{table*}[!t]
\caption{\textbf{Infidelity comparison.} Computed values show the average Infidelity over the 100 sample subsets. Results show differences between the different methods when applied to time series data. No category has shown a superior performance, although the gradient-based approaches were slightly better.}
\label{tab:infidelity}
\centering
\scalebox{0.75}{
\begin{tabular}{l|c|c|c|c|c|c}
\textbf{Method} & \textbf{Anomaly} & \textbf{CharacterTrajectories} & \textbf{ECG5000} & \textbf{FaceDetection} & \textbf{FordA} & \textbf{UWaveGestureLibraryAll} \\
\hline
\textbf{Gradient-based} & & & & & \\
GradientShap        & $2.3803$ & $\bm{1.1408}$ & $\bm{0.7897}$ & $0.0014$ & $1.3734$ & $11.4717$ \\
GuidedBackprop      & $2.4057$ & $1.1665$ & $0.8060$ & $0.0014$ & $1.3782$ & $11.6886$ \\
InputXGradient      & $\bm{2.3056}$ & $\bm{1.1475}$ & $0.8135$ & $0.0014$ & $1.3854$ & $11.5830$ \\
IntegratedGradients & $2.3594$ & $1.2064$ & $0.8260$ & $\bm{0.0013}$ & $\bm{1.3537}$ & $\bm{11.3763}$ \\
Saliency & $2.3788$ & $\bm{1.0921}$ & $0.8174$ & $0.0014$ & $\bm{1.3636}$ & $11.7546$ \\
\hline
\textbf{Perturbation-based} & & & & & \\
Dynamask            & $2.4382$ & $1.2650$ & $0.8271$ & $\bm{0.0013}$ & $1.3806$ & $11.6034$ \\
FeatureAblation     & $2.3859$ & $1.1513$ & $0.8459$ & $0.0014$ & $1.3869$ & $11.5511$ \\
FeaturePermutation  & $2.4015$ & $1.1654$ & $\bm{0.7949}$ & $0.0014$ & $1.3991$ & $11.5112$ \\
Occlusion           & $\bm{2.3430}$ & $1.2078$ & $0.8107$ & $0.0014$ & $1.3752$ & $\bm{11.3569}$ \\
\hline
\textbf{Miscellaneous} & & & & & \\
KernelShap          & $2.4115$ & $1.1802$ & $0.8288$ & $0.0014$ & $1.3785$ & $11.6568$ \\
Lime                & $2.4259$ & $1.1584$ & $\bm{0.8040}$ & $0.0014$ & $\bm{1.3732}$ & $11.6323$ \\
ShapleyValueSampling & $\bm{2.3352}$ & $1.1671$ & $0.8153$ & $0.0014$ & $1.3745$ & $\bm{11.4625}$ \\
\end{tabular}
}
\end{table*}

\begin{table*}[!t]
\caption{\textbf{Sensitivity comparison.} Computed values show the Sensitivity of a sample. Results show larger values for Lime and Shap-based approaches. Overall the performance of the perturbation-based approaches was superior to most of the other approaches.}
\label{tab:sensitivity}
\centering
\scalebox{0.75}{
\begin{tabular}{l|c|c|c|c|c|c}
\textbf{Method} & \textbf{Anomaly} & \textbf{CharacterTrajectories} & \textbf{ECG5000} & \textbf{FaceDetection} & \textbf{FordA} & \textbf{UWaveGestureLibraryAll} \\
\hline
\textbf{Gradient-based} & & & & & \\
GradientShap        & $0.9364$ & $0.6610$ & $0.9149$ & $0.9764$ & $1.0369$ & $1.0347$ \\
GuidedBackprop      & $0.1324$ & $0.1531$ & $0.0562$ & $0.1339$ & $\bm{0.0398}$ & $0.2057$ \\
InputXGradient      & $0.1890$ & $0.1017$ & $0.0709$ & $0.0952$ & $0.0924$ & $0.1927$ \\
IntegratedGradients & $0.1166$ & $0.1144$ & $0.0458$ & $\bm{0.0419}$ & $0.0906$ & $0.2086$ \\
Saliency            & $0.1902$ & $0.1126$ & $0.1841$ & $0.0995$ & $0.0762$ & $0.2220$ \\
\hline
\textbf{Perturbation-based} & & & & & \\
Dynamask            & $\bm{0.0000}$ & $\bm{0.0000}$ & $\bm{0.0000}$ & $\bm{0.0000}$ & $\bm{0.0000}$ & $\bm{0.0000}$ \\
FeatureAblation     & $\bm{0.0414}$ & $\bm{0.0360}$ & $\bm{0.0350}$ & $0.0581$ & $0.0463$ & $\bm{0.0444}$ \\
FeaturePermutation  & $\bm{0.0414}$ & $\bm{0.0360}$ & $\bm{0.0350}$ & $0.0581$ & $0.0463$ & $\bm{0.0444}$ \\
Occlusion           & $0.0645$ & $\bm{0.0167}$ & $\bm{0.0305}$ & $\bm{0.0506}$ & $\bm{0.0254}$ & $\bm{0.0352}$ \\
\hline
\textbf{Miscellaneous} & & & & & \\
KernelShap          & $1.0908$ & $0.9405$ & $0.2162$ & $0.9248$ & $0.8876$ & $1.0283$ \\
Lime                & $0.8221$ & $0.4986$ & $0.1408$ & $1.5613$ & $0.6974$ & $0.6378$ \\
ShapleyValueSampling & $0.9132$ & $0.3917$ & $0.1852$ & $0.5938$ & $0.5536$ & $0.3458$ \\
\end{tabular}
}
\end{table*}

\subsection{Infidelity \& Sensitivity}
The Infidelity measurements provide information about the change concerning the predictor function when perturbations to the input are applied. The metric derives from the completeness property of well-known attribution methods and is used to evaluate the quality of an attribution method. In the results in Table~\ref{tab:infidelity} the Infidelity represents a mean error using 100 perturbed samples for each approach. A lower Infidelity value corresponds to a better attribution method, and the optimal Infidelity value should be zero. The results show that the tested methods do differ by a large margin of less than $7.2\%$ on average, and in addition, the Infidelity values strongly depend on the dataset. Neither the gradient-based approaches nor the perturbation-based or other approaches are superior. The mean increase of the worst-performing and the best method was $7.2\%$. The experiments identified the highest increases for the CharacterTrajectories dataset ($15.8\%$) and the lowest for the FordA ($3.4\%$).

Further, the Sensitivity of the methods for a single sample was compared. Computationally, the Sensitivity is much more expensive but provides a good idea about the change in the attribution when the input is perturbed. Using the Sensitivity the robustness against of the methods concerning noise was evaluated. Ultimately, an attribution method tends to show low Sensitivity, although this depends on the model itself. In Table~\ref{tab:sensitivity} the results of the Sensitivity for all methods are presented. The results show that Dynamask has a Sensitivity of zero. Dynamask by design forces the importance values to be either one or zero. Although this is a benefit concerning the Sensitivity it results in a drawback when ranking the features as shown in the accuracy drop experiment. In addition, perturbation-based approaches have shown  $30.9\%$ better results on average concerning their Sensitivity across all datasets. The FordA dataset has shown the most significant difference between the attribution methods ($42.1\%$), while the CharacterTrajectories dataset has shown the lowest ($26.1\%$). Besides, the impressive performance of Dynamask, the Occlusion, FeatureAblation, and FeaturePermutation have shown results underlining their robustness against permutations. 

\subsection{Runtime}
\begin{figure}[!t]
\centering
\includegraphics[width=\linewidth]{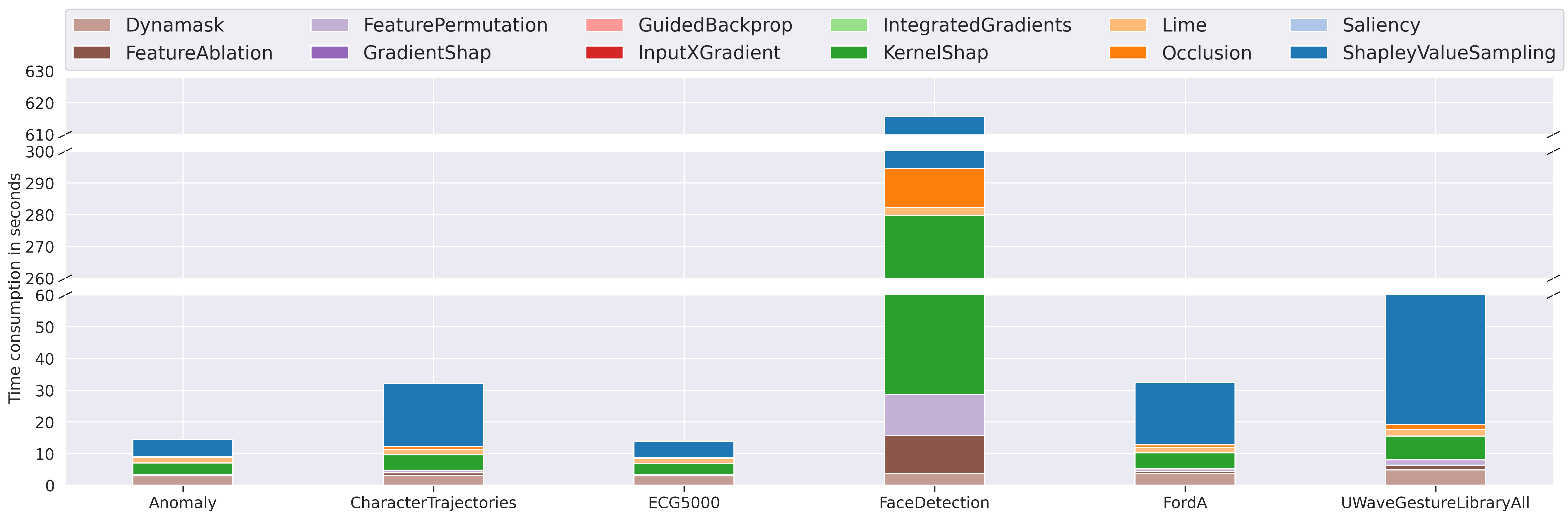}
\caption{\textbf{Time comparison.} Shows the time spend to compute the attribution of a single sample. Note that some bars are not visible due to their fast computation time compared to the other methods and the time of Dynamask is lowered by parameter optimization due to the otherwise unsuitable time consumption. Hardware: Quad-Core Intel Xeon processor, Nvidia GeForce GTX 1080 Ti, and 64 GB memory.}
\label{fig:time_consumption}
\end{figure}

The runtime and resource consumption are important aspects. Even though, the availability of resources increases, they are not unlimited. 
Depending on the throughput of the approach real-time interpretability can be possible. For mobile devices, the computation capacity is limited, and low resource dependencies are beneficial. A Quad-Core Intel Xeon processor, Nvidia GeForce GTX 1080 Ti, and 64 GB memory were used to compare the methods concerning their computational effort. The attribution and execution time for a single sample of each dataset was computed. In Figure~\ref{fig:time_consumption} shows that especially the simple gradient-based methods like the Saliency, IntegradtedGradients, and InputXGradient show a low computation time. On the other side methods like KernelShap and ShapleyValueSampling have shown increased time consumption. 
There is always the trade-off between how many samples are processed and the computational costs using SVS and KernelShap. During The analysis, the default values suggested in the corresponding papers of the methods were used. In the case of the FaceDetection dataset, the computational overhead of the FeatureAblation, FeaturePermutation, and Occlusion increased a lot as they strongly depend on the number of features. The FaceDetection dataset needs $41$ times longer than the anomaly dataset. Overall the computation time of the FaceDeteciton dataset is four times longer than the aggregated computation of all others. The characteristics of the FaceDetection dataset favor methods that are independent of the number of features. The high number of channels and time-steps when every data feature gets evaluated separately increases up to an unacceptable point. In addition, it has to be mentioned that only $100$ epochs instead of the default $1,000$ for each optimization of Dynamask were used to lower the computation times. The results show that this does not change the overall results of Dynamask but lowers the computational time by a factor of ten. Using the default 1000 epochs would not be suitable in any case as the computation time would increase by a factor of ten.

\subsection{Attribution Correlation}

\begin{figure}[!t]
\centering
\subfloat[CharacterTrajectories]{
\includegraphics[width=.96\linewidth]{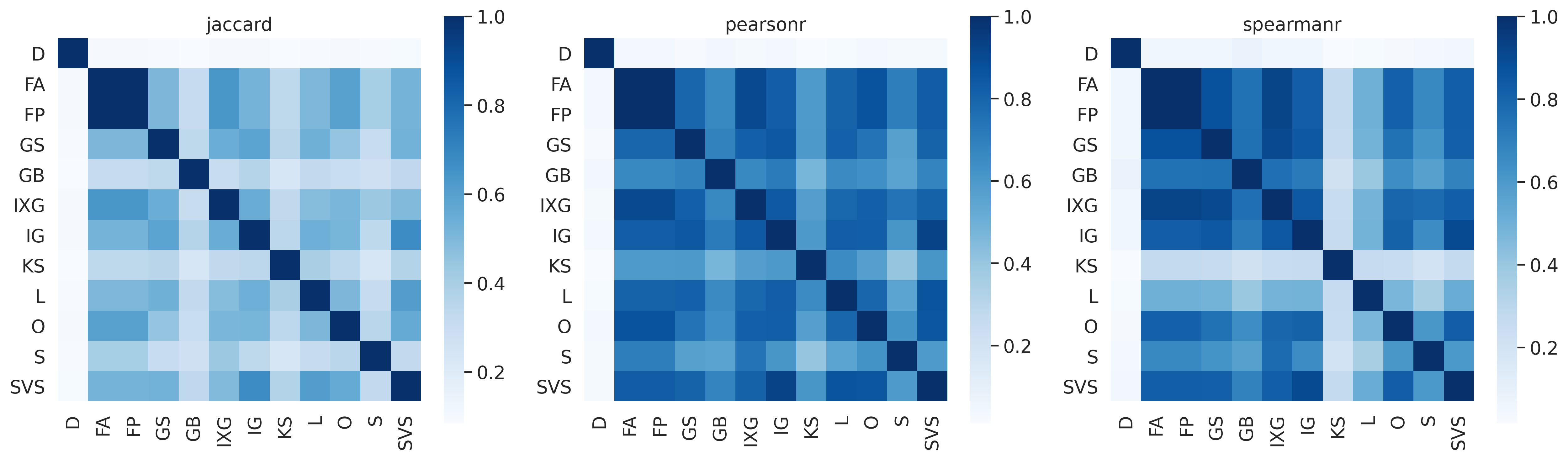}
\label{fig:Correlation_matrix_CharacterTrajectories}
}
\hfil
\subfloat[FordA]{
\includegraphics[width=.96\linewidth]{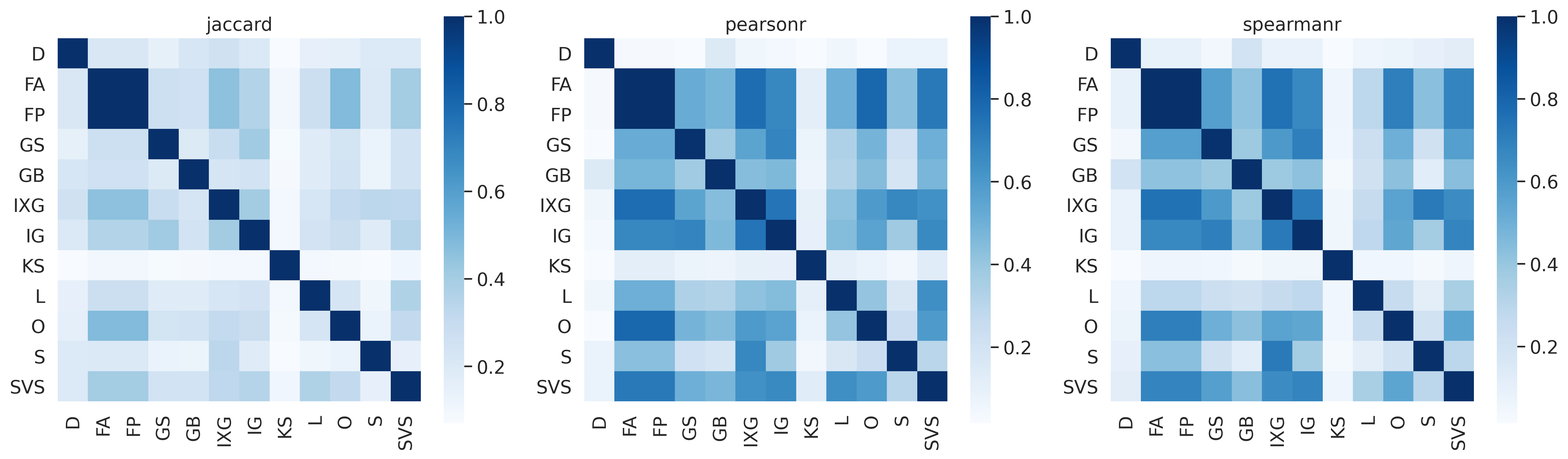}
\label{fig:Correlation_matrix_FordA}
}
\hfil
\subfloat[FaceDetection]{
\includegraphics[width=.96\linewidth]{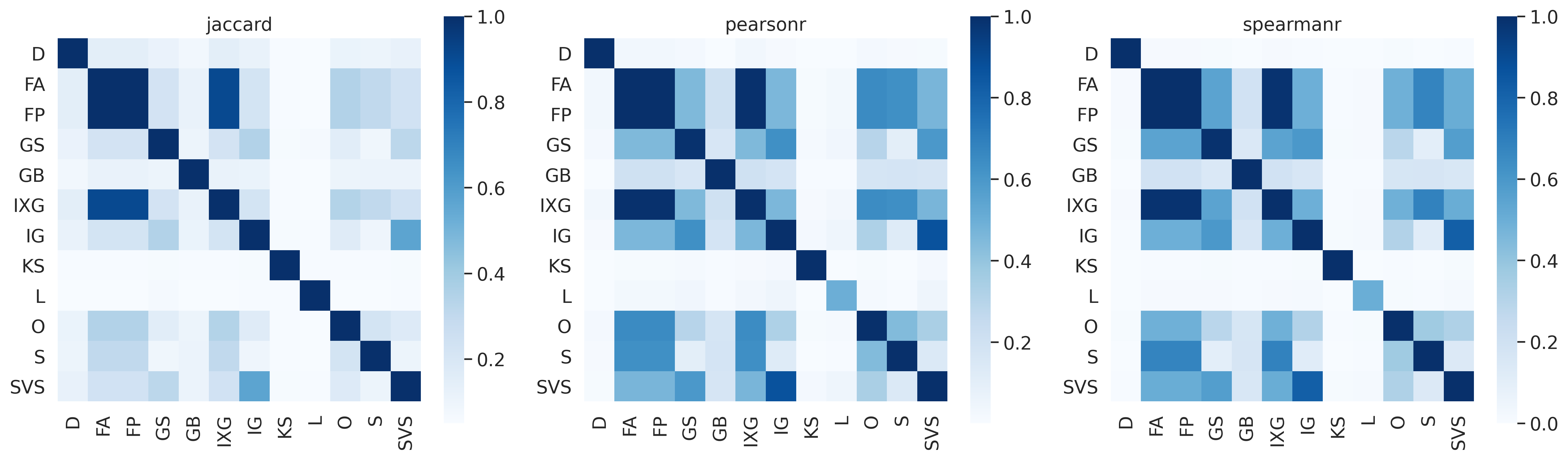}
\label{fig:Correlation_matrix_FaceDetection}
}
\caption{\textbf{Attribution correlation} Shows the average correlation/similarity of over $100$ attributions. The ten percent most important features were selected for the Jaccard similarity. The method names are shortened using only the capital characters. KernelShap shows a significantly lower correlation to other methods compared to all others. Feature Ablation and FeaturePermutation have shown a high correlation.}
\label{fig:Correlation_matrix}
\end{figure}

Another aspect is the correlation of the different attribution maps. Therefore, different correlation measurements were used, namely the Pearson correlation~\cite{benesty2009pearson}, Spearman correlation~\cite{myers2004spearman} and Jaccard Similarity~\cite{niwattanakul2013using}. The Pearson correlation measures the correlation between two series concerning their values. Spearman correlation is a ranked measurement that compares the ranks for each of the features. Finally, the Jaccard Similarity is used as a set-based measurement. During this experiment, the similarity of the attributions computed over the 100 test sample subsets was evaluated. Ultimately, only the important points matter concerning a correct attribution. That means intuitively, the similarity of the methods concerning irrelevant points. To consider that, percentile subsets of the important features were selected for the Jaccard Similarity to understand the agreement of the methods concerning those features. Summarizing the different similarity and correlation metrics, the absolute correlation using the Pearson correlation, the ranking using the Spearman correlation, and the important set of features using the Jaccard similarity were used. 

Figure~\ref{fig:Correlation_matrix} shows the results. The correlation matrices for the CharacterTrajectories and FordA dataset as the other datasets have similar show results. Overall every matrix shows the same behavior. FeatureAblation (FA) and FeaturePermutation (FP) are very similar. In addition, the Dynamask (D) approach and KernelShap (KS) are different from any of the others. This difference is the case for Dynamask, as the technique only makes a binary decision if a feature is significant or not. Intuitively, this should result in a high similarity for the Jaccard measurement. However, this is not the case as the attribution of Dynamask has an internal smoothing based on the loss used to optimize the mask. This smoothing will include less important features in the important feature set to preserve a continuous mask. Furthermore, Lime (L) and KernelShap (KS) seemed less similar to the other approaches.

\subsection{Dependency on Model Parameter}
Attribution methods should depend on the model parameter and the labels of the data. Therefore, the impact of label permutation and parameter randomization of the model was evaluated. The paper only shows the results using the CharacterTrajectories dataset as the results on the other datasets are similar.

\begin{figure}[!t]
\centering
\includegraphics[width=\linewidth]{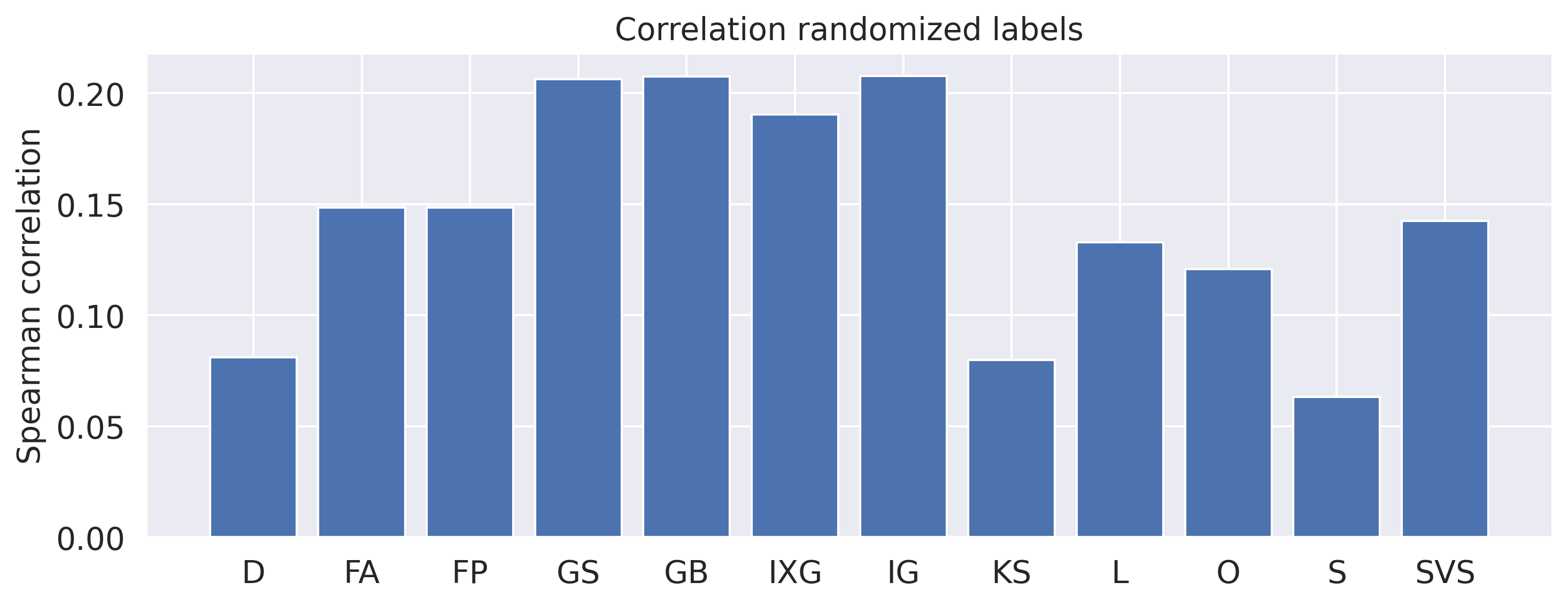}
\caption{\textbf{Attribution comparison.} Shows the Spearman correlation (rank correlation) of the attribution methods evaluated on the same model architecture using randomized training labels using the CharacterTrajectories dataset. The method names are shortened using only the capital characters. Dynamask, KernelShap, and Saliency show a significantly lower dataset dependence.}
\label{fig:correlation_data}
\end{figure}

The idea of the label permutation is that attribution methods should depend heavily on the labels. Good results in this experiment show a high intrinsic data characteristic dependence which is not a desired feature of an attribution method. The models were trained similar to the baseline model on the same training data but permuted the labels. This permutation results in a model that does not generalize well but learns to replicate the training set. In addition, this approach did not require the validation dataset. The accuracies of those models are very high for the training set. Nevertheless, they fail on the test set. Precisely speaking, these models do not have a label dependence. All models reached a near-perfect performance on the training set. Figure~\ref{fig:correlation_data} highlights that the correlation drops down to values between 0.05 and 0.2. Based on the overall low correlation, the attribution methods highly depend on the labels rather than dataset characteristics. GradientShap, GuidedBackprop, InputXGradient, and IntegratedGradients have shown three times larger correlations in contrast to Dynamak, KernelShap, and Saliency. However, their correlation is still low enough to justify the label dependency.

\begin{figure}[!t]
\centering
\includegraphics[width=\linewidth]{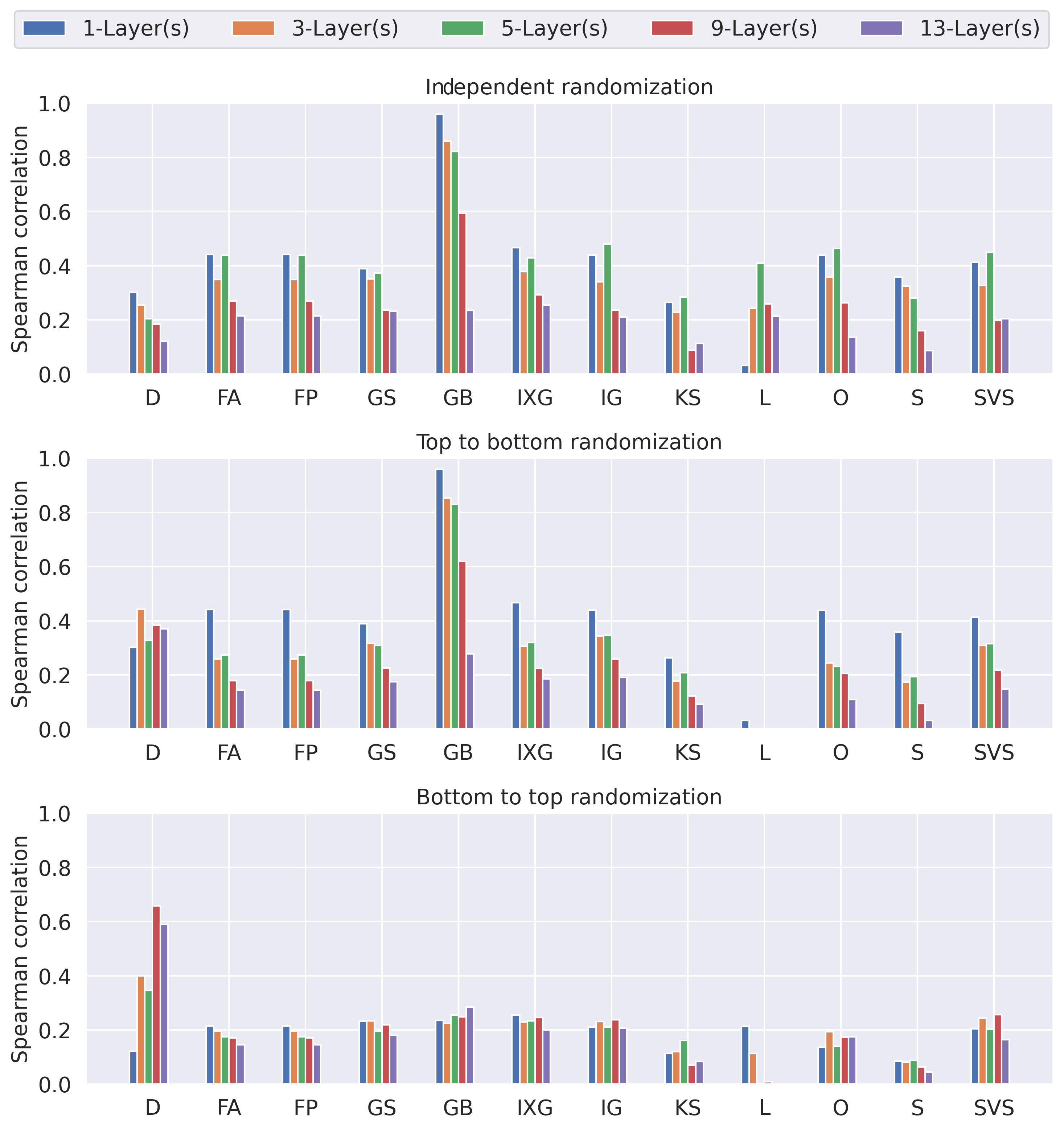}
\caption{\textbf{Correlation to original attribution.} Shows the Spearman correlation of the attribution methods evaluated on the trained model and randomized layer weights using the CharacterTrajectories dataset. Weights are either randomized for each layer independently, from top to the bottom layer or vice versa. Only layers with trainable parameters (conv, batchnorm, dense) are included when counting the number of randomized layers. The method names are shortened using only the capital characters. GuidedBackprop shows significant correlations when only the upper layers are randomized. The correlation of all other methods drops significantly.}
\label{fig:correlation_layer}
\end{figure}

In addition to the label permutation, layers of a correctly trained network were systematically randomized to understand the dependency concerning the model parameters. To understand the impact of the layers, each layer was randomized independently. Further, the model was randomized starting from the bottom to the top and vice-versa. The results in Figure~\ref{fig:correlation_layer} show all three approaches. Interestingly, the correlation of GuidedBackprop stays high when randomizing the top layers but significantly drops when randomizing the bottom layers. Randomizing the upper layers, the correlation of Guidedbackprop is close to the original attribution map, whereas the correlation of the other methods drops by $0.5$ or more. That suggests that this method is more based on the values of the first few layers. In addition, the results show that for all attribution techniques, a single randomized layer is enough to get an attribution that is no longer related to the original attribution map. This high dependency on the model parameter is the desired property. The top to bottom randomization further shows that except for the Dynamsk approach, the correlation continuously gets smaller when randomizing more layers. Finally, the bottom to top randomization highlights that the randomization of the first layer of the network is enough to produce attribution maps that are not related to the original.

\begin{figure*}[!t]
\centering
\includegraphics[width=\linewidth]{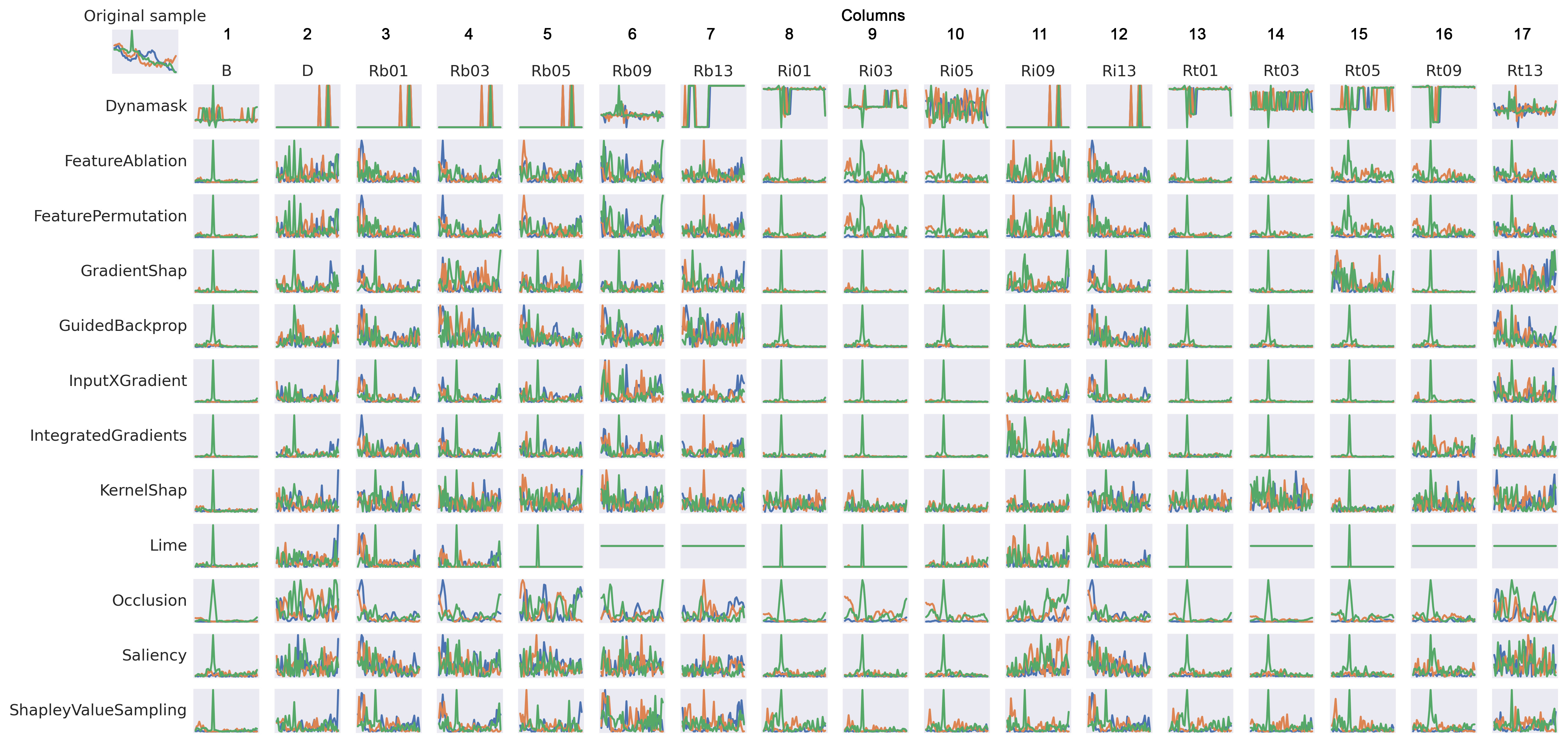}
\caption{\textbf{Visual comparison.} Shows all attributions for a selected anomaly sample. The important part is the peak of the sample. 'Ri', 'Rb', 'Rt', 'D', and 'B' correspond to the independent, bottom to top, top to bottom randomization, label randomization, and original attribution map. Only conv, batchnorm, and dense layers are counted. Changing the data labels during training significantly worsens the performance of all approaches except IntegratedGradients for the anomaly dataset. Overall randomizing lower layers resulted in much more noise compared to randomization in the upper layers.}
\label{fig:attribution_grid}
\end{figure*}

\subsection{Visual Attribution Comparison}
Figure~\ref{fig:attribution_grid} shows all computed attribution maps for a reference sample. Due to interpretability reasons, an anomalous instance of the anomaly dataset was selected. The example in the top left corner contains a single anomaly in one channel that is important for the classification. The rest of the figure shows the different attribution maps and the impact of randomization on the methods. The figure shows the robustness to randomized parameters. In the second column, the Integrated gradients approach was able to find the peak. This column corresponds to a model trained on randomized labels. Therefore, the model used in column two is not generalized and learned only to map the training data. Columns three to seven show a model randomization starting from the bottom layers. The results show that some methods still perform well when only one or three layers starting from the bottom are randomized other attribution methods directly collapsed. Columns eight to twelve show the independent layer randomization. Except for Dynamask, the attribution techniques were able to deal with up to handle the layer randomization in the upper layer of the network quite well, whereas all attribution methods collapsed when the lower layers were randomized. Columns thirteen to seventeen show the randomization starting from the top of the network. Most attribution methods were able to recover from the randomization for a high number of randomized layers. Overall the randomization of the lower layers changed the attribution much more concerning the noise. Interestingly, changes in the upper layers did not affect the attribution methods that much. 

\subsection{Continuity}
One aspect that is missing most times is attribution continuity. In the image domain, the use of superpixels solves this problem. However, in the time series domain, it is not that easy. Most of the attribution methods do not consider groups of values. In Table~\ref{tab:continuity} shows the evaluation of the continuity. The continuity calculates the absolute difference between the attribution value of a point $t$ and $t+1$ for each time-step and each channel. Using the mean across a sample provides a value that indicates how continuous the explanation is. Lower values correspond to an explanation that does not contain many switches from important to not important features. This measurement was computed over the 100 attributed samples and took the mean for each dataset. The results indicate that the perturbation-based approaches favor continuous explanations. Gradient-based methods overall have shown the worst performance. One reason for this is the noisy gradients used to compute the attribution maps.

\begin{table*}[!t]
\caption{\textbf{Continuity comparison.} Computed values show the mean continuity of the attribution maps. Lower values correspond to continuous maps. Continuity was calculated by shifting the attribution map, subtracting if from the original one, taking the absolute values, and computing the mean. Lower values are better. Perturbation-based methods have been shown to outperform gradient-based with respect to the continuity on almost all datasets. Specifically, Dynamask and Occlusion have been shown to perform well across all datasets.}
\label{tab:continuity}
\centering
\scalebox{0.8}{
\begin{tabular}{l|c|c|c|c|c|c}
\textbf{Method} & \textbf{Anomaly} & \textbf{CharacterTrajectories} & \textbf{ECG5000} & \textbf{FaceDetection} & \textbf{FordA} & \textbf{UWaveGestureLibraryAll} \\
\hline
\textbf{Gradient-based} & & & & & \\
GradientShap & $0.0947$ & $0.0368$ & $0.0616$ & $0.0613$ & $0.0813$ & $0.0543$ \\
GuidedBackprop & $0.1201$ & $0.0537$ & $0.0913$ & $0.0957$ & $\bm{0.0801}$ & $\bm{0.0526}$ \\
InputXGradient & $0.0801$ & $0.0390$ & $0.0508$ & $0.0620$ & $0.0855$ & $0.0537$ \\
IntegratedGradients & $0.0864$ & $0.0369$ & $0.0609$ & $0.0632$ & $0.0858$ & $0.0508$ \\
Saliency & $0.1176$ & $0.0748$ & $0.1439$ & $0.1170$ & $0.1229$ & $0.0842$ \\
\hline
\textbf{Perturbation-based} & & & & & \\
Dynamask & $\bm{0.0282}$ & $\bm{0.0014}$ & $\bm{0.0252}$ & $\bm{0.0107}$ & $\bm{0.0159}$ & $\bm{0.0015}$ \\
FeatureAblation & $0.0784$ & $0.0395$ & $0.0584$ & $0.0624$ & $0.0815$ & $0.0601$ \\
FeaturePermutation & $0.0784$ & $0.0395$ & $0.0584$ & $0.0624$ & $0.0815$ & $0.0601$ \\
Occlusion & $\bm{0.0623}$ & $\bm{0.0183}$ & $\bm{0.0419}$ & $\bm{0.0367}$ & $\bm{0.0535}$ & $\bm{0.0284}$ \\
\hline
\textbf{Miscellaneous} & & & & & \\
KernelShap & $0.1423$ & $0.1086$ & $0.0641$ & $0.1671$ & $0.1973$ & $0.1795$ \\
Lime & $0.1122$ & $0.0496$ & $\bm{0.0498}$ & $\bm{0.0010}$ & $0.0883$ & $0.0928$ \\
ShapleyValueSampling & $\bm{0.0773}$ & $\bm{0.0365}$ & $0.0505$ & $0.0583$ & $0.0885$ & $0.0713$ \\
\end{tabular}
}
\end{table*}

\section{\uppercase{Discussion}}
\label{sec:discussion}
A summarization and discussion in a detailed manner is offered to provide on choosing an attribution method. The different aspects and application scenarios are described below. First, it has to be mentioned that every attribution method has shown satisfying results. However, the choice of an attribution method should depend on the required characteristics. The overall results are presented in Table~\ref{tab:overall}. Teh resuls highlight that choosing an attribution method can be very important, as mentioned by Vermeire et al.~\cite{vermeire2021choose}.

Starting with the accuracy drop, the evaluation shows to which extend the methods rank the most and least significant features based on the impact on the accuracy. Most of the methods were able to show high-quality results across all datasets. However, there were some outstanding performances. Specifically, the perturbation-based were able to perform slightly better than the other methods on some datasets. Saliency and Dynamask have shown some weaknesses for some datasets, such as the CharacterTrajectories and FordA. Both methods require further adjustments and knowledge about the data to achieve good results. One example is the ratio of significant points for the Dynamask approach to select the correct number of features. If additional information is available, such as the ratio of selected features, methods like Dynamask can express their full potential. The attribution agreement shows similar results.

\begin{table}[!t]
\caption{\textbf{Overall Evaluation.} Overall results with respect to the different aspects evaluated in this paper. A = Accuracy Impact / Agreement, I = Infidelity, S = Sensitivity, R = Runtime, Ld = Label dependency, Md = Model Parameter Dependency, C = Continuity.}
\label{tab:overall}
\centering
\scalebox{0.8}{
\begin{tabular}{l|c|c|c|c|c|c|c}
\textbf{Method} & \textbf{A} & \textbf{I} & \textbf{S} & \textbf{R} & \textbf{Ld} & \textbf{Md} & \textbf{C}\\
\hline
\textbf{Gradient-based} & & & & & & & \\
GradientShap        & & $\oplus$ & & $\oplus$ & & & \\
GuidedBackprop      & $\oplus$ & & & $\oplus$ & & $\ominus$ & \\
InputXGradient      & & $\oplus$ & & $\oplus$ & & & \\
IntegratedGradients & & $\oplus$ & & $\oplus$ & & & \\
Saliency            & $\ominus$ & $\oplus$ & & $\oplus$ & $\oplus$ & & \\
\hline
\textbf{Perturbation-based} & & & & & & & \\
Dynamask            & $\ominus$ & & $\oplus$ & $\ominus$ & $\oplus$ & & $\oplus$ \\
FeatureAblation     & $\oplus$ & & $\oplus$ & $\ominus$ & & & \\
FeaturePermutation  & $\oplus$ & & $\oplus$ & $\ominus$ & & & \\
Occlusion           & & $\oplus$ & $\oplus$ & $\ominus$ & & & $\oplus$ \\
\hline
\textbf{Miscellaneous} & & & & & & & \\
KernelShap          & $\ominus$ & & & $\ominus$ & $\oplus$ & & \\
Lime                & & $\oplus$ & & $\oplus$ & & $\oplus$ & $\oplus$ \\
ShapleyValueSampling & $\oplus$ & & & $\ominus$ & & & $\oplus$ \\
\end{tabular}
}
\end{table}

Concerning Infidelity and Sensitivity, every method performed well, and no approach suffered more. The results show that gradient-based methods obtained the best Infidelity results. It was the opposite for the Sensitivity. Especially, GradientShap, InputXGradient, and Saliency approaches are robust against significant perturbations in the input space (Infidelity). On the other side, the Dynamask, FeaturePermutation, and Occlusion approaches have shown good robustness concerning changes in the attribution when small perturbations to the input are applied (Sensitivity). Dynamask has a loss that forces a binary decision whether a feature is selected or not ensures this behavior. Using attribution methods with low Sensitivity values in cases where adversarial attacks can occur is suggested.

The runtime aspect gets critical when the use case requires near real-time explanations.  In addition, the results have shown that the dataset characteristics are relevant. The findings show that approaches based on the sequence length and number of channels suffer from very high runtimes for single samples. These runtimes make it impossible to use them in a real-time scenario. However, if the time consumption is not of interest, this aspect is not relevant. Furthermore, gradient-based methods are less dependent on the dataset characteristics and very suitable when time matters. Contrarily, besides Dynamask and Lime, the perturbation-based approaches suffer from the number of features. In the case of Lime, the number of samples required to populate the space to train the surrogate model increases with a higher number of features. Dynamask does not suffer from the feature number.  However, the approach needs an additional training phase. This training requires multiple epochs and in addition repetitions based on the different areas checked during the training. Ultimately, the backpropagation needs resources and time. Based on the computational times, the use of ShapelyValueSampling and KernelShap in real-time scenarios is nearly impossible. For completeness, it has to be mentioned that it is possible to tweak hyperparameters.

The label permutation and layer randomization provided insights concerning the role of the model parameters during the attribution computation. Intuitively, all methods have shown a high dependency on the labels of the data. Training a model with randomized targets has shown, the attributions depend on the labels as they should. Although all methods have shown this dependency, the Saliency, Dynamask, KernelShap, and Lime have shown more dependence on the targets. Concerning the model parameters, the results show that randomizing any layer results in changes of the attribution maps. Besides, the GuidedBackprop attribution maps significantly change after any modification. Specifically, Lime collapses completely. This collapse emphasizes that Lime directly depends on the model, and GuidedBackprop is relying more on data. An explanation for this behavior is that some methods detect dataset differences. Especially in the image domain, it was shown that some attribution methods can act like an edge detectors. 

Finally, continuity plays a pivotal role in human understanding. In use cases that include human evaluation, it is beneficial to have continuous attribution maps. Imagine there is a significant frame with many important but some less important features. It might be superior to mark the whole window as important, although this covers some insignificant features. In the time series domain, the context matters, and continuous attribution maps are easier to understand. The results show that the Dynamask approach, Lime, Occlusion, and ShapleyValueSampling are superior concerning their continuity. Intuitively, the attribution maps produced by gradient-based techniques look noisy, whereas permutation-based look smoother. Dynamask includes a loss term that ensures a smoother attribution map. Lime and ShapleyValueSampling produce smoother maps. The results suggest using a perturbation-based approach if a human inspection is relevant.

Comparing the gradient-based, perturbation-based, and other approaches, every category has shown advantages over the other category in some aspects. Generally, gradient-based methods are fast, show high Infidelity, label dependency but are noisy, not continuous, and suffer concerning the Sensitivity. In contrast to gradient-based methods, perturbation-based approaches produce continuous maps, shine concerning the Sensitivity, label dependency but suffer when it comes to the runtime.

\section{\uppercase{Conclusion}}
A comprehensive evaluation of a large set of state-of-the-art attribution methods applicable to time series was performed. The results show that most attribution methods can identify significant features without prior knowledge about the data. In the evaluation, the perturbation-based approaches have shown slightly superior performance in the data occlusion game. In addition, the results are validated by measuring the agreement of the methods using different correlation and similarity measurements. Except for Dynamask and KernelShap, the correlation between the attribution methods showed high values. Further experiments were conducted to highlight the high dependence of the attribution methods on the model and the target labels. Only Guided-Backprop has shown lower reliance on the top layers of the network. Concerning Infidelity, the gradient-based attribution methods showed superior performance. The perturbation-based attribution methods are superb concerning Sensitivity and continuity. Continuity is an important aspect when it comes to human interpretability. The results hold across a set of different tasks, sequence lengths, feature channels, and the number of samples. Furthermore, the results show that the choice of an attribution method depends on the target scenario, and different aspects like runtime, accuracy, continuity, noise are indispensable.

\section*{\uppercase{Acknowledgment}}
This work was supported by the BMBF projects SensAI (BMBF Grant 01IW20007) and the ExplAINN (BMBF Grant 01IS19074). We thank all members of the Deep Learning Competence Center at the DFKI for their comments and support.

\bibliographystyle{apalike}
{\small \bibliography{bibliography}}

\begin{thebibliography}{}

\bibitem[Abdul et~al., 2020]{abdul2020cogam}
Abdul, A., von~der Weth, C., Kankanhalli, M., and Lim, B.~Y. (2020).
\newblock Cogam: Measuring and moderating cognitive load in machine learning
  model explanations.
\newblock In {\em Proceedings of the 2020 CHI Conference on Human Factors in
  Computing Systems}, pages 1--14.

\bibitem[Adebayo et~al., 2018]{adebayo2018sanity}
Adebayo, J., Gilmer, J., Muelly, M., Goodfellow, I., Hardt, M., and Kim, B.
  (2018).
\newblock Sanity checks for saliency maps.
\newblock {\em arXiv preprint arXiv:1810.03292}.

\bibitem[Allam and Dhunny, 2019]{allam2019big}
Allam, Z. and Dhunny, Z.~A. (2019).
\newblock On big data, artificial intelligence and smart cities.
\newblock {\em Cities}, 89:80--91.

\bibitem[Ancona et~al., 2017]{ancona2017towards}
Ancona, M., Ceolini, E., {\"O}ztireli, C., and Gross, M. (2017).
\newblock Towards better understanding of gradient-based attribution methods
  for deep neural networks.
\newblock {\em arXiv preprint arXiv:1711.06104}.

\bibitem[Ancona et~al., 2019]{ancona2019gradient}
Ancona, M., Ceolini, E., {\"O}ztireli, C., and Gross, M. (2019).
\newblock Gradient-based attribution methods.
\newblock In {\em Explainable AI: Interpreting, Explaining and Visualizing Deep
  Learning}, pages 169--191. Springer.

\bibitem[Bagnall et~al., 2021]{tsc2021datasets}
Bagnall, A., Lines, J., Vickers, W., and Keogh, E. (2021).
\newblock The uea \& ucr time series classification repository.

\bibitem[Benesty et~al., 2009]{benesty2009pearson}
Benesty, J., Chen, J., Huang, Y., and Cohen, I. (2009).
\newblock Pearson correlation coefficient.
\newblock In {\em Noise reduction in speech processing}, pages 1--4. Springer.

\bibitem[Bibal et~al., 2020]{bibal2020impact}
Bibal, A., Lognoul, M., de~Streel, A., and Fr{\'e}nay, B. (2020).
\newblock Impact of legal requirements on explainability in machine learning.
\newblock {\em arXiv preprint arXiv:2007.05479}.

\bibitem[Crabbé and van~der Schaar, 2021]{Crabbe2021Dynamask}
Crabbé, J. and van~der Schaar, M. (2021).
\newblock Explaining time series predictions with dynamic masks.
\newblock In {\em Proceedings of the 38-th International Conference on Machine
  Learning (ICML 2021)}. PMLR.

\bibitem[Das and Rad, 2020]{das2020opportunities}
Das, A. and Rad, P. (2020).
\newblock Opportunities and challenges in explainable artificial intelligence
  (xai): A survey.
\newblock {\em arXiv preprint arXiv:2006.11371}.

\bibitem[Do{\v{s}}ilovi{\'c} et~al., 2018]{dovsilovic2018explainable}
Do{\v{s}}ilovi{\'c}, F.~K., Br{\v{c}}i{\'c}, M., and Hlupi{\'c}, N. (2018).
\newblock Explainable artificial intelligence: A survey.
\newblock In {\em 2018 41st International convention on information and
  communication technology, electronics and microelectronics (MIPRO)}, pages
  0210--0215. IEEE.

\bibitem[Fisher et~al., 2019]{fisher2019all}
Fisher, A., Rudin, C., and Dominici, F. (2019).
\newblock All models are wrong, but many are useful: Learning a variable's
  importance by studying an entire class of prediction models simultaneously.
\newblock {\em J. Mach. Learn. Res.}, 20(177):1--81.

\bibitem[Huber et~al., 2021]{huber2021benchmarking}
Huber, T., Limmer, B., and Andr{\'e}, E. (2021).
\newblock Benchmarking perturbation-based saliency maps for explaining deep
  reinforcement learning agents.
\newblock {\em arXiv preprint arXiv:2101.07312}.

\bibitem[Ivanovs et~al., 2021]{ivanovs2021perturbation}
Ivanovs, M., Kadikis, R., and Ozols, K. (2021).
\newblock Perturbation-based methods for explaining deep neural networks: A
  survey.
\newblock {\em Pattern Recognition Letters}.

\bibitem[Karliuk, 2018]{karliuk2018ethical}
Karliuk, M. (2018).
\newblock Ethical and legal issues in artificial intelligence.
\newblock {\em International and Social Impacts of Artificial Intelligence
  Technologies, Working Paper}, (44).

\bibitem[Lundberg and Lee, 2017]{lundberg2017unified}
Lundberg, S.~M. and Lee, S.-I. (2017).
\newblock A unified approach to interpreting model predictions.
\newblock In {\em Proceedings of the 31st international conference on neural
  information processing systems}, pages 4768--4777.

\bibitem[Mitchell et~al., 2021]{mitchell2021sampling}
Mitchell, R., Cooper, J., Frank, E., and Holmes, G. (2021).
\newblock Sampling permutations for shapley value estimation.
\newblock {\em arXiv preprint arXiv:2104.12199}.

\bibitem[Myers and Sirois, 2004]{myers2004spearman}
Myers, L. and Sirois, M.~J. (2004).
\newblock Spearman correlation coefficients, differences between.
\newblock {\em Encyclopedia of statistical sciences}, 12.

\bibitem[Nielsen et~al., 2021]{nielsen2021robust}
Nielsen, I.~E., Rasool, G., Dera, D., Bouaynaya, N., and Ramachandran, R.~P.
  (2021).
\newblock Robust explainability: A tutorial on gradient-based attribution
  methods for deep neural networks.
\newblock {\em arXiv preprint arXiv:2107.11400}.

\bibitem[Niwattanakul et~al., 2013]{niwattanakul2013using}
Niwattanakul, S., Singthongchai, J., Naenudorn, E., and Wanapu, S. (2013).
\newblock Using of jaccard coefficient for keywords similarity.
\newblock In {\em Proceedings of the international multiconference of engineers
  and computer scientists}, volume~1, pages 380--384.

\bibitem[Perc et~al., 2019]{perc2019social}
Perc, M., Ozer, M., and Hojnik, J. (2019).
\newblock Social and juristic challenges of artificial intelligence.
\newblock {\em Palgrave Communications}, 5(1):1--7.

\bibitem[Peres et~al., 2020]{peres2020industrial}
Peres, R.~S., Jia, X., Lee, J., Sun, K., Colombo, A.~W., and Barata, J. (2020).
\newblock Industrial artificial intelligence in industry 4.0-systematic review,
  challenges and outlook.
\newblock {\em IEEE Access}, 8:220121--220139.

\bibitem[Ribeiro et~al., 2016]{lime}
Ribeiro, M.~T., Singh, S., and Guestrin, C. (2016).
\newblock "why should {I} trust you?": Explaining the predictions of any
  classifier.
\newblock In {\em Proceedings of the 22nd {ACM} {SIGKDD} International
  Conference on Knowledge Discovery and Data Mining, San Francisco, CA, USA,
  August 13-17, 2016}, pages 1135--1144.

\bibitem[Shrikumar et~al., 2016]{shrikumar2016not}
Shrikumar, A., Greenside, P., Shcherbina, A., and Kundaje, A. (2016).
\newblock Not just a black box: Learning important features through propagating
  activation differences.
\newblock {\em arXiv preprint arXiv:1605.01713}.

\bibitem[Siddiqui et~al., 2019]{siddiqui2019tsviz}
Siddiqui, S.~A., Mercier, D., Munir, M., Dengel, A., and Ahmed, S. (2019).
\newblock Tsviz: Demystification of deep learning models for time-series
  analysis.
\newblock {\em IEEE Access}, 7:67027--67040.

\bibitem[Simonyan et~al., 2013]{simonyan2013deep}
Simonyan, K., Vedaldi, A., and Zisserman, A. (2013).
\newblock Deep inside convolutional networks: Visualising image classification
  models and saliency maps.
\newblock {\em arXiv preprint arXiv:1312.6034}.

\bibitem[Springenberg et~al., 2014]{springenberg2014striving}
Springenberg, J.~T., Dosovitskiy, A., Brox, T., and Riedmiller, M. (2014).
\newblock Striving for simplicity: The all convolutional net.
\newblock {\em arXiv preprint arXiv:1412.6806}.

\bibitem[Sundararajan et~al., 2017]{sundararajan2017axiomatic}
Sundararajan, M., Taly, A., and Yan, Q. (2017).
\newblock Axiomatic attribution for deep networks.
\newblock In {\em International Conference on Machine Learning}, pages
  3319--3328. PMLR.

\bibitem[Vermeire et~al., 2021]{vermeire2021choose}
Vermeire, T., Laugel, T., Renard, X., Martens, D., and Detyniecki, M. (2021).
\newblock How to choose an explainability method? towards a methodical
  implementation of xai in practice.
\newblock {\em arXiv preprint arXiv:2107.04427}.

\bibitem[Yeh et~al., 2019]{yeh2019fidelity}
Yeh, C.-K., Hsieh, C.-Y., Suggala, A., Inouye, D.~I., and Ravikumar, P.~K.
  (2019).
\newblock On the (in) fidelity and sensitivity of explanations.
\newblock {\em Advances in Neural Information Processing Systems},
  32:10967--10978.

\bibitem[Zeiler and Fergus, 2014]{zeiler2014visualizing}
Zeiler, M.~D. and Fergus, R. (2014).
\newblock Visualizing and understanding convolutional networks.
\newblock In {\em European conference on computer vision}, pages 818--833.
  Springer.

\bibitem[Zhang and Zhu, 2018]{zhang2018visual}
Zhang, Q. and Zhu, S.-C. (2018).
\newblock Visual interpretability for deep learning: a survey.
\newblock {\em arXiv preprint arXiv:1802.00614}.

\end{thebibliography}

\end{document}